\documentclass[lettersize,journal]{IEEEtran}
\usepackage[utf8]{inputenc}
\usepackage{amsmath,amsfonts}
\usepackage{array}
\usepackage[caption=false,font=normalsize,labelfont=sf,textfont=sf]{subfig}
\usepackage{textcomp}
\usepackage{stfloats}
\usepackage{verbatim}
\usepackage{graphicx}
\usepackage{cite}
\usepackage{overpic}
\usepackage{amsfonts}
\usepackage{multirow}
\usepackage{amssymb}
\usepackage{float}
\usepackage{bm}
\usepackage{bbding}
\usepackage{pifont}
\usepackage{hyperref}
\usepackage{booktabs}
\usepackage{color}
\usepackage{colortbl}
\usepackage[numbers,sort&compress]{natbib}
\definecolor{mygray}{gray}{.9}

\usepackage{algorithm}
\usepackage{algorithmic}

\begin{document}
\title{Causality-based Cross-Modal Representation Learning for Vision-and-Language Navigation}
\author{
Liuyi~Wang,~Zongtao~He,~Ronghao~Dang,~Huiyi Chen,~Chengju~Liu,~Qijun~Chen,~\IEEEmembership{Senior~Member,~IEEE}

\IEEEcompsocitemizethanks{
\IEEEcompsocthanksitem L. Wang, Z. He, R. Dang, C. Liu and Q. Chen are with the School of Electronic and Information Engineering, Tongji University, China. C. Liu is also with the Tongji Artificial Intelligence (Suzhou) Research Institute and Frontiers Science Center for Intelligent Autonomous Systems. E-mail: \{wly, xingchen327, dangronghao, liuchengju, qjchen\}@tongji.edu.cn.
\IEEEcompsocthanksitem H. Chen is with Rutgers University, New Brunswick, USA. E-mail: hc991@rutgers.edu.
\IEEEcompsocthanksitem Corresponding authors: Chengju Liu, Qijun Chen
}}

\markboth{Journal of \LaTeX\ Class Files,~Vol.~14, No.~8, August~2015}%
{Shell \MakeLowercase{\textit{et al.}}: Bare Demo of IEEEtran.cls for Computer Society Journals}

\maketitle

\begin{abstract}
Vision-and-Language Navigation (VLN) has gained significant research interest in recent years due to its potential applications in real-world scenarios. However, existing VLN methods struggle with the issue of spurious associations, resulting in poor generalization with a significant performance gap between seen and unseen environments. In this paper, we tackle this challenge by proposing a unified framework CausalVLN based on the causal learning paradigm to train a robust navigator capable of learning unbiased feature representations. Specifically, we establish reasonable assumptions about confounders for vision and language in VLN using the structured causal model (SCM). Building upon this, we propose an iterative backdoor-based representation learning (IBRL) method that allows for the adaptive and effective intervention on confounders. Furthermore, we introduce the visual and linguistic backdoor causal encoders to enable unbiased feature expression for multi-modalities during training and validation, enhancing the agent's capability to generalize across different environments. Experiments on three VLN datasets (R2R, RxR, and REVERIE) showcase the superiority of our proposed method over previous state-of-the-art approaches. Moreover, detailed visualization analysis demonstrates the effectiveness of CausalVLN in significantly narrowing down the performance gap between seen and unseen environments, underscoring its strong generalization capability.
\end{abstract}


\section{Introduction}
\label{sec_introduction}
\IEEEPARstart{E}{mbodied} Artificial Intelligence (AI) is a field dedicated to developing agents, such as robots, that learn to creatively solve tasks by actively interacting with and navigating through the dynamic environment. One fundamental task in this field is Vision-and-Language Navigation (VLN)~\cite{anderson2018vision,qi2020reverie,ku2020room,wang2021visual,lin2021adversarial,xie2022vision}, where an agent is tasked with navigating through an environment based on visual observations and natural language instructions. VLN has gained significant research interest in recent years due to its potential applications in real-world scenarios, such as personal assistants and in-home robots.
VLN is a challenging task that requires the seamless integration of computer vision (CV), natural language processing (NLP), and robotics. While various studies have proposed solutions in model construction and learning strategies, leading to improved navigation performance on curated datasets, deploying VLN in complex and ever-changing environments still presents significant challenges. One key obstacle is the substantial performance gap between navigation performance in seen and unseen environments due to the scarcity of the human-annotated dataset~\cite{fu2020counterfactual} and the bias hidden in the dataset~\cite{zhang2021diagnosing}. As a result, the practicality and reliability of VLN systems in real-world applications are limited.

Previous studies have primarily focused on addressing the data scarcity issue in VLN by employing data augmentation. One common approach is to use a speaker-follower structure~\cite{fried2018speaker,tan2019learning,wang2022counterfactual,wang2023res}. In this approach, a speaker model generates pseudo instructions to provide additional training data for the follower.
Furthermore, image augmentation methods such as environmental dropout~\cite{tan2019learning} and GAN-based approaches~\cite{li2022envedit} have been proposed to simulate novel visual environments. Some methods propose random mixing of labeled data~\cite{jain2019stay,liu2021vision} or collecting data from the web~\cite{majumdar2020improving,guhur2021airbert}. Moreover, pre-training methods~\cite{hao2020towards,hong2021vln,qiao2022hop,qiao2023hop_plus} on large-scale datasets have also shown promising results by broadening the inductive bias. Although these methods have made progress in improving generalization through data augmentation, such approaches may be insufficient due to the inherent dataset biases~\cite{peters2017elements,yang2021deconfounded}. The substantial spurious correlation may be useful for in-domain learning but harmful for out-of-domain testing~\cite{zhang2020devlbert}. Therefore, it becomes vital to learn the true cause-and-effect relationship of variables, enabling the model to grasp the underlying logic of the task and thus improve its adaptability to unknown data distributions, such as unseen scenes and instructions.

\begin{figure}[tb]
    \centering
    \includegraphics[scale=0.6]{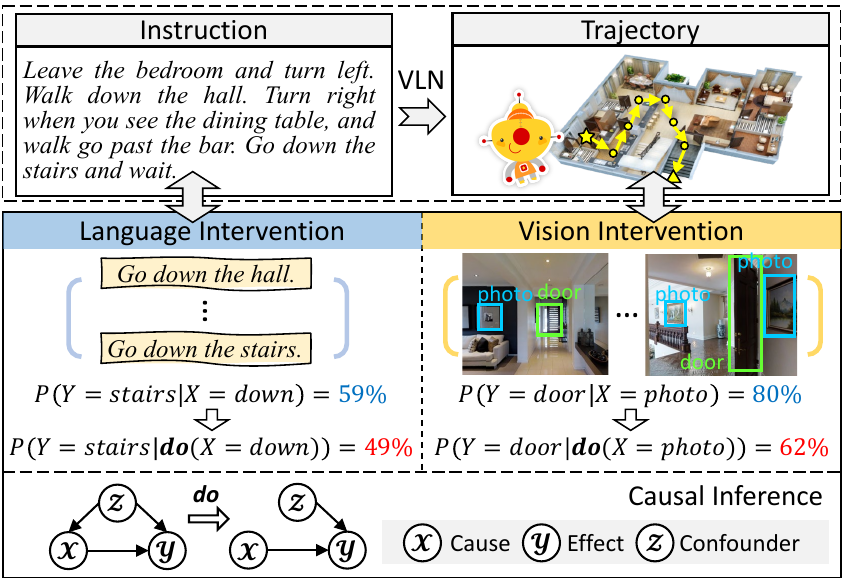}
    \caption{Vision-and-Language Navigation (VLN) involves an agent navigating through visual environments based on language instructions. In this paper, we leverage intervention on both language and vision modalities to learn unbiased features and enhance the generalization of the model.}
    \label{fig_top}
\end{figure}
In recent years, there has been a growing interest in applying causal learning approaches~\cite{pearl2018book} to vision-and-language tasks to address the issue of biased perception and inference caused by spurious correlation~\cite{wang2020visual,qi2020two,zhang2020devlbert,yang2021deconfounded}. To explore the presence of bias in VLN, we conduct a toy experiment~\cite{wang2020visual} on the Room-to-Room (R2R) dataset~\cite{anderson2018vision}. As shown in Fig.~\ref{fig_top}, the conditional probability $P(door|photo)$ (the probability of a \texttt{door} appearing given the presence of a \texttt{photo}) is relatively high at $80\%$. However, there are no such strong causal relationships between these objects. To mitigate this bias issue, a backdoor adjustment method is proposed by~\cite{pearl2018book}, utilizing a \textit{do} operator to control the influence of potential confounders during outcome assessment (also known as \textit{intervention}). Here, confounders $\mathcal{Z}$ represent variables that impact both the inputs $\mathcal{X}$ and outcomes $\mathcal{Y}$. The backdoor adjustment is a powerful technique to break the link between confounders and inputs, leading to true causal reasoning. By doing this, $P(door|do(photo))$ declines to $62\%$. 
Consequently, the original inference based on associations is transformed into a causality-based inference.

Building upon this line of research, we pose the following question: \textit{Can we capture and model the underlying causal relationships in VLN and generate unbiased feature representations that enhance the robustness of navigational agents?} By shifting our focus from dataset collection to the design of unbiased models~\cite{dang2022unbiased}, we can gain a deeper understanding of the intrinsic characteristics of VLN and generate more meaningful representations. However, the application of existing causal learning methods to VLN presents significant challenges that demand careful consideration. Firstly, the partially observable Markov decision process of VLN~\cite{hong2021vln} introduces complexities that hinder the direct pursuit of causal inference from input to output. Secondly, the presence of confounders both within vision and language further complicates the establishment of a robust perceptual and decision-making network.

This paper presents a pioneering effort to introduce causal learning to VLN, in order to enhance the generalization capabilities of navigators. We propose a novel framework called CausalVLN to facilitate this endeavor. First, We use the structural causal model (SCM) to represent the causal relationships in VLN. Next, we introduce an iterative backdoor-based representation learning (IBRL) method to enable intervention on confounders within the network. Unlike previous approaches that directly intervene from inputs to outputs, our method focuses on the causal inference between inputs and features, allowing for more flexible and adaptive interventions. Additionally, we propose an adaptive and updated confounder feature dictionary during training to prevent substantial representation gaps, unlike previous approaches that employ a fixed confounder feature dictionary~\cite{wang2020visual,yang2021deconfounded}. Then, it is essential to identify the confounders in a causal learning system~\cite{pearl2018book}. Therefore, we make assumptions about observable confounders present in vision and language modalities, respectively. Based on the characteristics of VLN, the \textit{objects} and \textit{room types} are considered to be observable confounders for vision, while \textit{landmark} and \textit{direction} words are chosen for language. The visual and linguistic backdoor causal encoders are proposed to realize the intervention of these confounders. Finally, we utilize the memory-augmented global-local cross-modal fusion module from our previous work DSRG~\cite{wang2023dual} to enable the agent to align and leverage features from different modalities, capturing valuable historical cues throughout the navigation.

The experimental results on three widely adopted VLN datasets, namely R2R\cite{anderson2018vision}, REVERIE\cite{qi2020reverie}, and RxR~\cite{ku2020room}, demonstrate the superiority of CausalVLN compared to previous methods. Furthermore, the quantitative and qualitative evaluations provide compelling evidence for the effectiveness of the assumed observable confounders and the proposed IBRL module. These advancements significantly improve navigation performance and help bridge the performance gap between seen and unseen settings.

\begin{figure*}[t]
    \centering
    \includegraphics[scale=0.73]{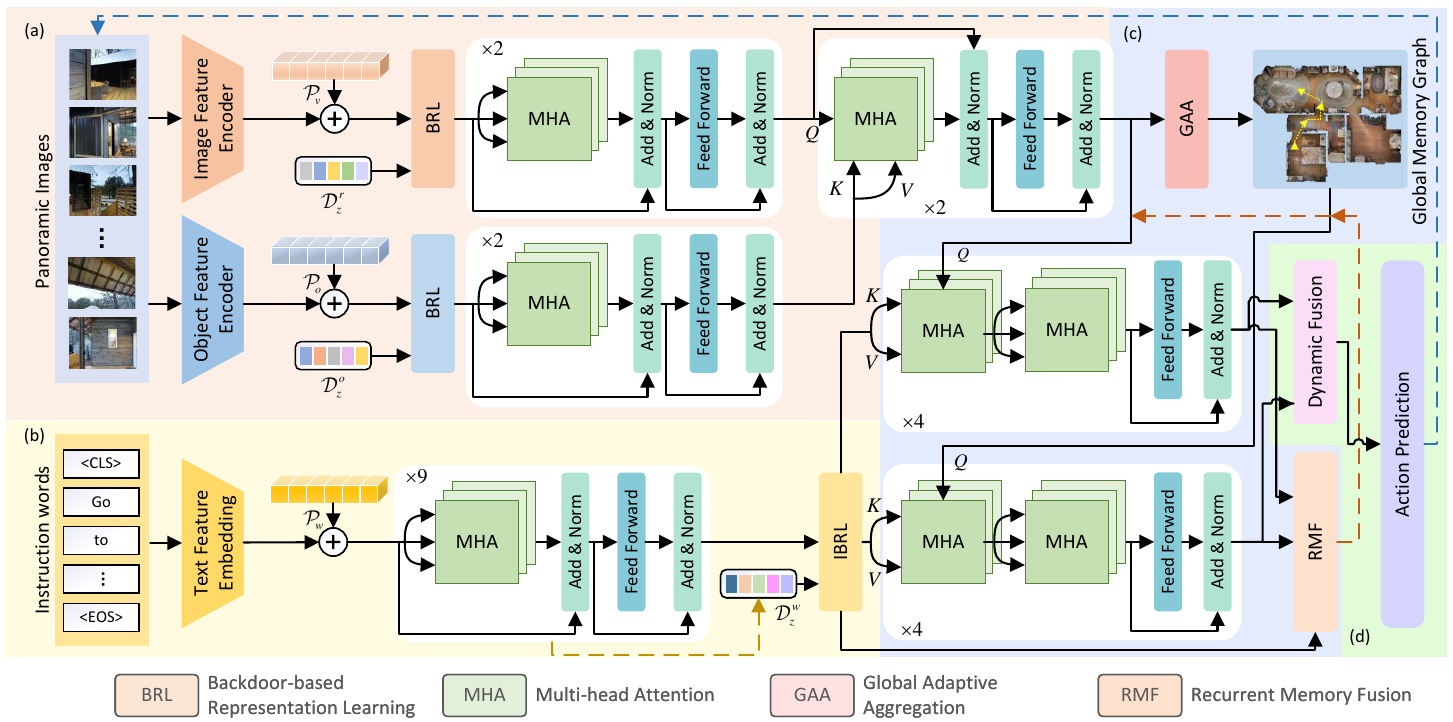}
    \caption{The overview of the proposed CausalVLN model. The visual backdoor causal encoder (a) and linguistic backdoor causal encoder (b) are used to learn the causality-based features for vision and language, respectively. The iterative update strategy for backdoor-based representation learning is employed in the language branch due to the participation of BERT in end-to-end training. The memory-augmented global-local cross-modal fusion (c) and dynamic action prediction (d) are used to enhance long-term navigation and adaptive decision-making.}
    \label{fig_overview}
\end{figure*}
In summary, our main contributions are as follows:
\begin{itemize}
    \item We introduce a novel CausalVLN framework, which represents the first attempt to incorporate causal learning into vision-and-language navigation (VLN), leading to the development of robust navigators.
    \item We emphasize the causal inference between inputs and features, proposing an iterative backdoor-based representation learning (IBRL) method that enables adaptive interventions within the network during training.
    \item We make reasonable assumptions about the vision and language inputs in VLN. Direction-landmark tokens are identified as confounders for instructions, leading to the proposal of a linguistic backdoor causal encoder. Similarly, objects and room types are considered confounders for visual observations, resulting in the introduction of a visual backdoor causal encoder.
    \item Through comprehensive experiments on three popular VLN datasets--R2R, REVERIE, and RxR, we demonstrate the significant advancements achieved by the proposed CausalVLN model. 
\end{itemize}

\section{Related Work}
\label{sec_related_work}

\subsection{Vision-and-Language Navigation}
\label{subsec_vln}
The Vision-and-Language Navigation (VLN) task~\cite{anderson2018vision}, which involves guiding an agent through a real environment using natural language instructions, has gained significant attention in recent years. In early works, the models were based on the encoder-decoder framework~\cite{wang2020vision,an2021neighbor,dang2022unbiased,he2023mlanet}. Upon further investigation, researchers discovered the training limitations of the small-scale R2R dataset. To overcome this, the speaker-follower models~\cite{fried2018speaker,wang2022counterfactual,dou2022foam,wang2023pasts,wang2023res} were proposed to improve generalization using an independent speaker model to generate pseudo instructions based on back translation~\cite{tan2019learning}. These developments opened up opportunities for larger models, such as transformers~\cite{vaswani2017attention}, to advance in the field of VLN. Hao~\emph{et al.}~\cite{hao2020towards} introduced PREVALENT, a transformer-based model that was pre-trained on a large-scale dataset generated by the speaker. Hong~\emph{et al.}~\cite{hong2021vln} proposed a recurrent VLN-BERT that maintained cross-modal states for the agent. 
Due to the powerful long-distance encoding capability, transformer-based models~\cite{chen2021history, chen2022think,qiao2023hop_plus} have shown strong performance. Further, Majumdar~\emph{et al.}~\cite{majumdar2020improving} and Guhur~\emph{et al.}~\cite{guhur2021airbert} collected a large number of image-text pairs from the web. REM~\cite{liu2021vision} randomly cross-connects the environments. EnvEdit~\cite{li2022envedit} created new environments by editing existing images. Kamath~\emph{et al.}~\cite{kamath2023new} used Marky~\cite{wang2022less} to generate visually-ground instructions.

In more recent works, some navigation methods have incorporated object information to enhance environment understanding~\cite{dang2022search,dang2023multiple}. ORIST~\cite{qi2021road} and SOAT~\cite{moudgil2021soat} proposed to concatenate object features with the scene features in transformer encoders. BiasVLN~\cite{zhang2021diagnosing} observed that the low-level image features cause the environmental bias. OAAM~\cite{qi2020object} processed textual object and action encoding and made them match the visual perception. NvEM~\cite{an2021neighbor} used soft attention to encode the neighboring objects of candidates. SEvol~\cite{chen2022reinforced} utilized the graph to construct relationships of objects. Our previous work DSRG~\cite{wang2023dual} proposed a dual semantic-augmented module to model the semantics explicitly. Although previous methods have highlighted the importance of data augmentation and the leverage of semantic cues, we argue that these approaches are necessary but insufficient. This is because they may inadvertently introduce dataset bias, which can result in misguided inferences when applied to new environments. Therefore, we propose using causal learning to intervene in the observable confounders present in the input. This approach enables the model to learn more precise, reliable, and unbiased representations.

\subsection{Causal Representation Learning}
\label{subsec_causal}
Causal inference~\cite{pearl2018book} is an emerging research direction in the field of machine learning. Unlike traditional approaches, it focuses on discovering high-level causal relationships from low-level data instead of merely observing correlations between variables. In recent years, several studies have applied causal learning to vision-and-language tasks~\cite{qi2020two,zhang2020devlbert,niu2021counterfactual,wang2023weakly}, such as image recognition~\cite{wang2021causal,zhang2022multiple,wang2023meta}, image captioning~\cite{yang2021deconfounded,liu2022show}, and visual question answering~\cite{niu2021counterfactual,li2022representation}. Specifically, Lopez-Paz~\emph{et al.}~\cite{lopez2017discovering} first proposed the observational causal discovery technique that targets the causal dispositions of objects in images. Wang~\emph{et al.} proposed the visual commonsense R-CNN~\cite{wang2020visual} that uses the normalized weighted geometric mean (NWGM)~\cite{baldi2014dropout} to approximate the softmax operation and thus employs the causal intervention to serve as an improved visual region classifier. Niu~\emph{et al.}~\cite{niu2021counterfactual} suggested a counterfactual inference framework to capture language bias as the direct causal effect of questions on answers. Liu~\emph{et al.}~\cite{liu2022show} proposed CIIC that applies the backdoor adjustment to construct the interventional object detector and transformer decoder.

Although the aforementioned methods have shown the effectiveness of causal learning in deep-learning models, applying causal learning to VLN remains a significant challenge. This is because the distinctive characteristics of the observed visual environment (\textit{e.g.}, navigating through indoor rooms in a first-person perspective) and the detailed step-by-step instructions in VLN diverge from other vision-and-language tasks, demanding meticulous contemplation of the underlying causal relationships. In addition, current methods aim to achieve direct causal intervention from input to output~\cite{wang2020visual, liu2022show}. However, this is challenging to implement in deep learning networks with intricate structures. Therefore, we narrow down our hypothesis to a causal learning process from input to feature. Furthermore, previous methods utilized a fixed pre-processed confounding dictionary for intervention, without considering the offset issue as the network weights were updated. To address these challenges, we establish a structured causal model based on the characteristics of VLN and propose a novel iterative causal representation learning method to enable features to fully express the essence of input.

\section{Methodology}
\label{sec_methology}
Fig.~\ref{fig_overview} provides an overview of the proposed model, which aims to improve the generalization of VLN agents by using causal learning to mitigate bias caused by confounders. We first provide a brief introduction to the VLN task in Sec.~\ref{subsec_task_formulation}. Next, we utilize the structural causal model to establish the causal graph for VLN in Sec.~\ref{subsec_scm}. Then, in Sec.~\ref{subsec_IBRL}, we propose an iterative backdoor-based representation learning method (IBRL) that is designed to capture the causal relationships between inputs and hidden features. In Sec.~\ref{subsec_VFCR} and Sec.~\ref{subsec_LFCR}, the visual and linguistic confounder dictionaries are identified and the encoders are created, respectively. In Sec.~\ref{subsec_memory-augmented}, the memory-augmented global-local cross-modal fusion module is adopted to aggregate modalities and boost historical dependency. The dynamic decision-making strategy in Sec.~\ref{subsec_decision-making_strategy} is used to predict the final prediction. Finally, we describe the pre-training and fine-tuning strategy in Sec.~\ref{subsec_training_strategy}. These components work together to enable effective and reliable decision-making for our proposed model.

\subsection{Task Formulation}
\label{subsec_task_formulation}
The Vision-and-Language Navigation (VLN) task involves training an embodied agent to navigate real indoor environments based on natural language instructions. The Matterport3D simulator~\cite{chang2017matterport3d} provides a graph-based environment $\mathcal{G}=\{P,\xi\}$, where $P$ and $\xi$ represent navigable nodes and connectivity edges, respectively. Formally, the agent receives two modalities: a natural language instruction $\mathcal{I}=\{w_1,w_2,...,w_L\}$, where $L$ is the length of the instruction, and a panorama observation of the current location represented as a set of image patches $\mathcal{V}=\{v_1,v_2,...,v_N\}$, where $N$ is the number of images and is set to 36. The agent also knows its orientation, represented as a 4D vector $\gamma = (\sin{\theta},\cos{\theta},\sin{\phi},\cos{\phi})$, where $\theta$ and $\phi$ denote the heading and elevation direction, respectively. During navigation, the agent receives visual observations of the current location and selects the next point from visible candidates. Upon reaching the target point, the agent sends a \textit{stop} signal. The navigation is considered successful if the stop location falls within a 3-meter radius of the ground-truth position.
\begin{figure}[tb]
    \centering
    \includegraphics[scale=0.74]{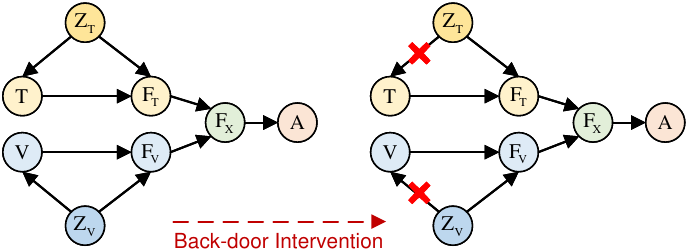}
    \caption{Illustration of the proposed causal graph. $V,T$ and $A$ denote the visual inputs, language inputs, and action prediction, respectively. $Z_v$ and $Z_T$ denote the confounder of the vision and the language. $F_V, F_T$ and $F_X$ are the hidden representations.}
    \label{fig_SCM}
\end{figure}

\subsection{Structural Causal Model for VLN}
\label{subsec_scm}
To make reasonable causal inferences, it is necessary to make explicit assumptions about the underlying causal structure of the task. One popular approach to modeling causality is the structural causal model (SCM)~\cite{glymour2016causal}, which represents the causal relationships between variables as a directed acyclic graph. The graph comprises nodes and edges, where nodes represent variables and edges represent causal relationships between them. For example, $X \rightarrow Y$ means that $X$ is the cause of $Y$. The confounder $Z$ is a variable that acts as a common cause for two other variables. As illustrated in Fig.~\ref{fig_SCM}, we use the SCM to formulate the causalities among the visual input $V$, linguistic input $T$, visual confounder $Z_V$, linguistic confounder $Z_T$, visual features $F_T$, linguistic features $F_V$, fused features $F_x$ and action prediction $A$. Conventional visual and linguistic encoders in VLN methods typically focus on learning the features of two links: $T\rightarrow F_T$ and $V\rightarrow F_V$. However, they overlook the presence of confounders and spurious correlations within these modalities. In contrast, our method takes a novel causal perspective by decoupling the confounders from the two modalities, providing fundamental solutions to address this issue. The rationale behind our proposed causal graph is elaborated in the following.

First, $Z_T/Z_V \rightarrow T/V$ means that, in practice, collecting the dataset can be influenced by various contextual factors, such as environmental and cognitive limitations, leading to a shift in the composition ratio. Additionally, when extracting features using a pre-trained model trained on a large dataset, relevant confounders present in the dataset can introduce bias into the learned features $F$, leading to $Z_T/Z_F\rightarrow F_T/F_V$. Such confounders create interference with the true causal relationship between the input and features, resulting in biased feature learning. This phenomenon is prevalent in both visual and textual feature extraction. As a result, the biased vision and language features can lead to biased fused features $F_x$, ultimately affecting the accuracy and generalization of action prediction in the open world.

In this paper, our objective is to train a navigator to acquire a deep understanding of the true causal relationship $T/V \rightarrow F_T/F_V$, ensuring that the navigator derives features from the inputs based on reasoning rather than relying on spurious correlations induced by confounders. To achieve this, we borrow the idea of the backdoor adjustment~\cite{pearl2000models,wang2020visual} to learn the causality, which will be discussed in the following section.

\subsection{Iterative Backdoor-Based Representation Learning}
\label{subsec_IBRL}
\begin{figure}[tb]
    \centering
    \includegraphics[width=\linewidth]{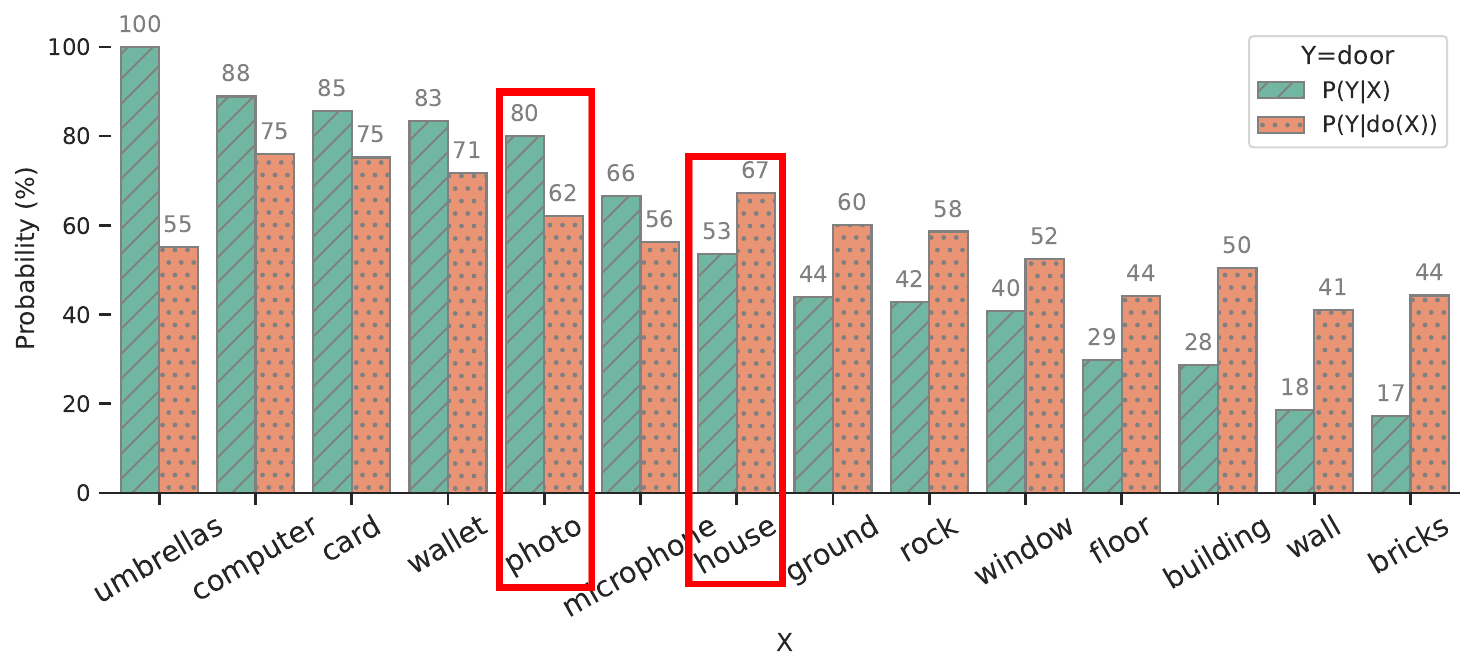}
    \includegraphics[width=\linewidth]{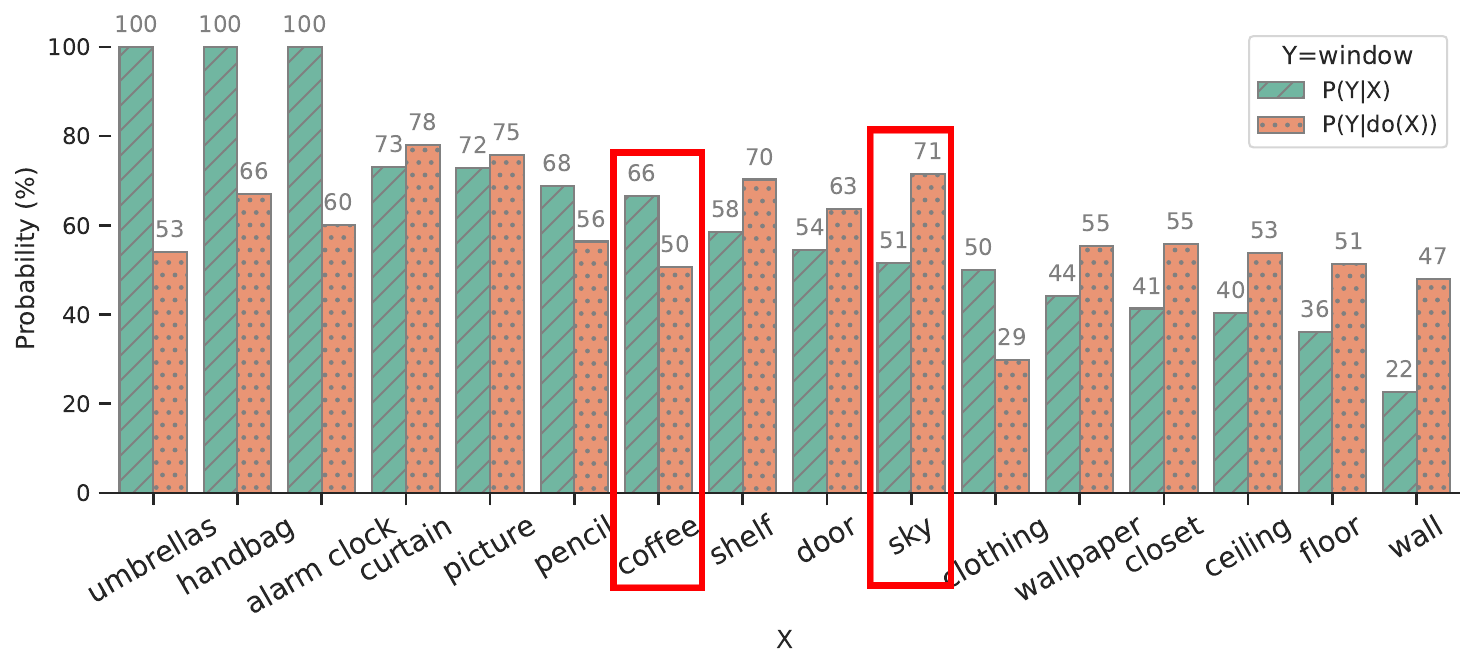}
    \caption{The statistics of $P(Y|X)$ and $P(Y|do(X))$. Only part of the object pairs is visualized to avoid clutter. With the help of the intervention, the causality of some pairs gets correction and becomes more commonsense.}
    \label{fig_vis_statistics}
\end{figure}

\subsubsection{Backdoor Adjustment} 
It is essential to distinguish between \textit{``observation"} and \textit{``control"} in causal learning, where the former involves passive observation of naturally occurring relationships, while the latter refers to active manipulation of the independent variable to establish causality. Based on Bayes' theorem, we can model the observational probability of classical classifiers as Eq.~(\ref{eq_observational_p}).
\begin{gather}
\label{eq_observational_p}
    P(Y|X)=\sum_{z\in Z} P(Y|X,z)P(z|X)
\end{gather}

Due to the existence of the shortcut raised by the confounder $Z$, the model tends to learn the spurious correlation induced by the backdoor path. To settle this problem, the backdoor adjustment is proposed to revise Eq.~(\ref{eq_observational_p}) to $P(Y|do(X))$, where \textit{do}-operation is a hypothetical intervention on a variable that controls for potential confounding. By doing this, the causal link between $Z$ and $X$ is cut off, rendering $Z$ and $X$ independent. To illustrate, let $P(Y|X)$ and $P_m(Y|X)$ denote the probabilities before and after treatment on the causal graph, respectively, such that $P(Y|do(X))=P_m(Y|X)$. According to the rules of invariance and independence, we can derive three additional equations: $P_m(z)=P(z)$, $P_m(Y|X,z)=P(Y|X,z)$ and $P_m(z|X)=P(z)$. By considering these equations together, we can now obtain:
\begin{align}
    P(Y|do(X)) &:= P_m(Y|X) \\
               &= \sum_{z\in Z} P_m(Y|X,z)P_m(z|X) \\
               &= \sum_{z\in Z} P_m(Y|X,z)P_m(z) \\
               &= \sum_{z\in Z} P(Y|X,z)P(z) \\
               &= \mathbb{E}_z(f_\theta(x,z)) \label{eq_backdoor}
\end{align}
where $f_\theta(\cdot)$ denotes the network output with the learnable parameters $\theta$. To demonstrate the effectiveness of the intervention, we conducted a toy experiment using objects in images from the R2R dataset, as depicted in Fig.~\ref{fig_vis_statistics}.
\begin{align}
    P(Y|X) &= \frac{P(X,Y)}{P(X)} \label{eq_toy_experiment1} \\
    P(Y|do(X)) &= \sum_z \frac{P(Y,X,z)P(z)}{P(X,z)} \label{eq_toy_experiment2}
\end{align}

In Fig.~\ref{fig_vis_statistics}, it illustrates the significant differences in likelihood before and after the intervention. For instance, the probability of $\texttt{house}\rightarrow \texttt{door}$ increases while $\texttt{photo} \rightarrow \texttt{door}$ decreases after the intervention. This shift aligns more with our intuitive understanding that houses typically have doors, while photos are not necessarily associated with doors. 
Notably, some 100\% associations (potentially due to the small size of these kinds of objects in the dataset) have also been reduced, resulting in associations between objects that better align with common sense reasoning. 
\begin{figure}[tb]
    \centering
    \includegraphics[scale=0.85]{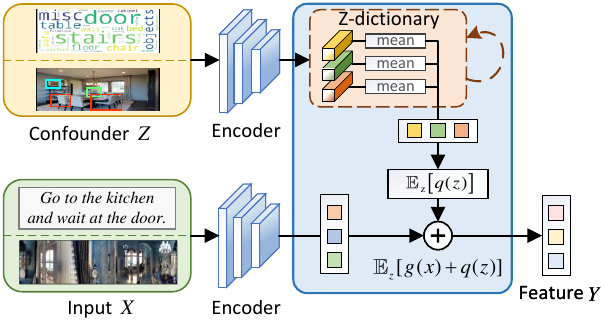}
    \caption{Illustration of the iterative backdoor-based representation learning.}
    \label{fig_IBCR}
\end{figure}
\subsubsection{Iterative Backdoor-Based Representation Learning Module}
Based on the above analysis, we aim to design a causal representation learning module within the network to enforce the learned features to faithfully reflect the inputs and minimize the interference caused by confounders. Previous methods~\cite{wang2020visual,liu2022show,yang2021deconfounded} attempted to establish the direct causal inference between inputs and outputs, using the normalized weighted geometric mean (NWGM) method~\cite{baldi2014dropout} to approximate the softmax function of the last layer of the network. However, we argue that this approach restricts the application of causal learning solely to the output layer, overlooking potential biases learned in the shallower layers. To address this limitation, we shift our focus to the causality between inputs and intermediate features. These methods enable us to more flexibly calculate Eq.~\eqref{eq_backdoor} and intervene within the network. 
Concretely, as shown in Fig.~\ref{fig_IBCR}, the linear function $f_\theta(\mathbf{x},\mathbf{z})=g(\mathbf{x})+q(\mathbf{z})$ is used to model the causal representation learning. According to the additive property of expectations, we have:
\begin{gather}
    \mathcal{E}_f = \mathbb{E}_z[f_\theta(\mathbf{x},\mathbf{z})] = g(\mathbf{x})+\mathbb{E}_z[q(\mathbf{z})] 
\end{gather}

Therefore, the main problem now relies on the calculation of $\mathcal{E}_z=\mathbb{E}_z[q(\mathbf{z})]$. To achieve the feature representation of confounders, we first apply the dictionary learning algorithm~\cite{wang2020visual} to construct a basis dictionary $\mathcal{D}_z=\{\mathbf{z}_1,\mathbf{z}_2,...,\mathbf{z}_K\}$ from the representations of all the training samples: 
\begin{align}
    \mathbf{z}_k = \frac{1}{N_k}\sum_n \mathbf{z}_{k,n} \label{eq_dz}
\end{align}
where $N_k$ and $K$ are the numbers of $k$-th type of the confounder and the total types of confounders, respectively. Then, we design two methods to present $\mathcal{E}_z$ as follows:

\textbf{Type 1: The statistic-based model.} The first method is to calculate the probability $p(z)$ based on statistical results:
\begin{align}
    p(z_k)=\frac{N_k}{\sum_j^K N_j} \label{eq_pz}
\end{align}

Then the expectation can be achieved by 
\begin{align}
E_z = \sum_k \mathbf{z}_k W_z p(z_k)
\end{align}
where $W_z\in\mathbb{R}^{d_m\times d_h}$ is the learnable parameter. The broadcast operation is then used to achieve $\mathcal{E}_{z1}=[E_z;...;E_z] \in \mathbb{R}^{L_x\times d_h}$ to allow the aggregation with input matrix $X \in \mathbb{R}^{L_x \times d_m}$ with the learnable parameter $W_x \in \mathbb{R}^{d_m\times d_h}$:
\begin{align}
    \mathcal{E}_f=XW_x + \mathcal{E}_{z1} \label{eq_Ef1}
\end{align}

\textbf{Type 2: The attention-based model.} The second method is to use the efficient query-key-value operation~\cite{liu2022show} to approximate $\mathcal{E}_{z2}$ by assigning $X\in \mathbb{R}^{L_x\times d_m}$ as query and $Z\in \mathbb{R}^{K \times d_m}$ as key and value:
\begin{align}
    \mathcal{E}_{z2} &= \textit{Softmax}(\frac{(XW_q)(ZW_k)^{T}}{\sqrt{d_k}})(ZW_v) \label{eq_attention_E1}\\
    \mathcal{E}_f &= XW_x + \mathcal{E}_{z2} \label{eq_attention_E2}
\end{align}
where $Z\in \mathbb{R}^{K\times d_m}$. It is not difficult to find that Eq.~\eqref{eq_attention_E1} and Eq.~\eqref{eq_attention_E2} are similar to the operations of the multi-head attention with a residual layer in the transformer~\cite{vaswani2017attention}, so the implementation is straightforward. The key distinction between Type-1 and Type-2 lies in the calculation of $\mathcal{E}_z$ and its contribution to various input $X$. In Type-1, $\mathcal{E}_z$ remains constant and has an equal effect on all input variables. However, in Type-2, different values of $\mathcal{E}_z$ are assigned to each input variable, thereby varying its impact.

In addition, in contrast to previous methods that use a static feature dictionary throughout the entire training process, we propose that the representation of confounders should also evolve to adjust to the fluctuations in network weights. Therefore, we suggest an iterative approach to update the dictionary. The comprehensive IBRL method is depicted in Algorithm~\ref{alg_IBCR}. To improve training efficiency, the image and object features undergo pre-processing using fixed extraction models~\cite{anderson2018bottom,radford2021learning}, while BERT~\cite{devlin2018bert} works as the language encoder and participates in the end-to-end training process. Therefore, we focus the iterative update on the causal learning module of the language encoder (see Fig.~\ref{fig_overview}). In the following, we will detail the implementation of the intervention on vision and language, respectively. 
\begin{algorithm}[tb]
\begin{algorithmic}[1]
\caption{Iterative Backdoor-Based Representation Learning (IBRL)}
\label{alg_IBCR}
\REQUIRE Dataset $\mathcal{S}$, update iterations $T$, learning rate $\alpha$.
\ENSURE Trained model $f_{\theta}$, confounder dictionary $\mathcal{D}_{z}$.
\STATE Initialize $\mathcal{D}_z$ on $\mathcal{S}$ with Eq.~\eqref{eq_dz};
\STATE Initialize $t = 0$;
\REPEAT
\STATE Sample a batch of data $(x, y) \sim \mathcal{S}$;
\STATE Compute $\mathcal{E}_f$ based on Eq.~\eqref{eq_Ef1} or Eq.~\eqref{eq_attention_E2} and get the causal feature $\widetilde{x}$;
\STATE Output the prediction $\hat{y}$ based on the causal feature $\widetilde{x}$;
\STATE Calculate the loss $\mathcal{L}(\hat{y}, y)$;
\IF{$\mathcal{L}$ is minimized or $t=T$}
\STATE Update the confounder dictionary $\mathcal{D}_z$:
\FOR{$x \in \mathcal{S}$}
    \STATE Extract $z$ from $x$;
    \STATE Compute $\widetilde{\mathbf{z}}=f_{\theta}^{e}(z)$ where $f_{\theta}^{e}$ is the encoder part;
\ENDFOR
\FOR{$\mathbf{z}_k \in \mathcal{D}_z$}
    \STATE Update $\mathbf{z}_k$: $\mathbf{z}_k \leftarrow \frac{1}{N_k} \sum_{n}\widetilde{\mathbf{z}}_{k,n}$;
\ENDFOR
\STATE Set $t = 0$;
\ELSE
\STATE Set $t = t + 1$;
\ENDIF
\STATE Update the network parameters: $\theta \leftarrow \theta - \alpha \nabla\theta \mathcal{L}(\hat{y}, y)$;
\UNTIL{convergence or maximum iterations is reached.}
\RETURN{$f_\theta$ and $\mathcal{D}_z$.}
\end{algorithmic}
\end{algorithm}

\begin{figure*}[t]
    \centering
    \includegraphics[scale=0.74]{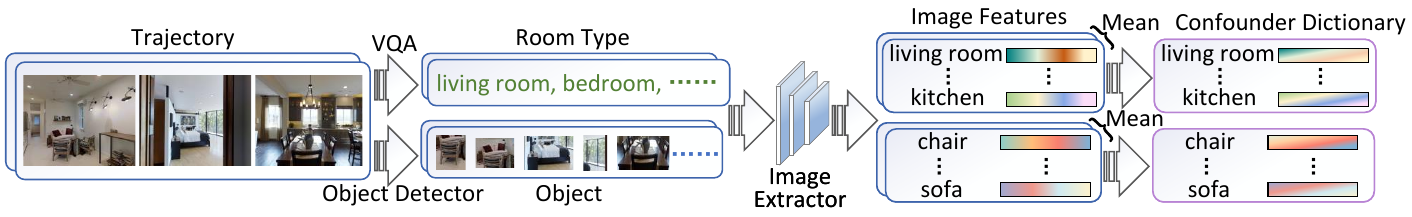}
    \caption{The process of constructing the confounder dictionary for images, considering objects and room types as confounders. Object classes and features are obtained using the bottom-up attention model~\cite{anderson2018bottom}, while room type labels are obtained by sending the prompt to the pre-trained VQA model BLIP~\cite{li2022blip}. The image features are extracted using CLIP~\cite{radford2021learning}, and the confounder dictionary is constructed by taking the average features for each item.}
    \label{fig_dict_traj}
\end{figure*}
\begin{figure*}[t]
    \centering
    \includegraphics[scale=0.74]{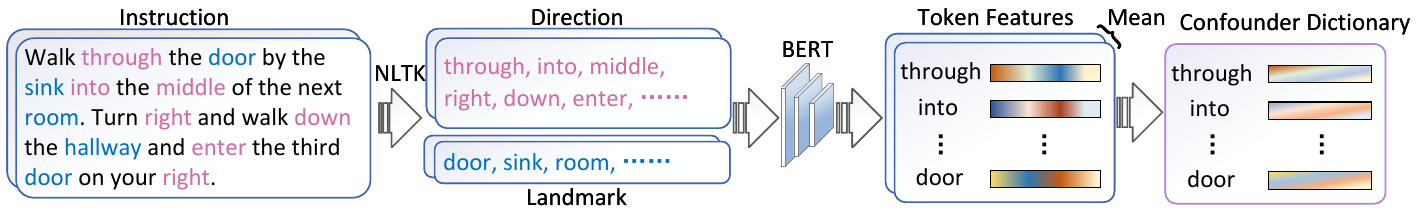}
    \caption{The confounder dictionary construction process for instructions, considering the direction and landmark tokens as confounders. The dictionary is built by taking the average token features extracted from different instructions using the pre-trained BERT~\cite{devlin2018bert}.}
    \label{fig_dict_instr}
\end{figure*}
\subsection{Visual Feature Causal Representation Learning} 
\label{subsec_VFCR}
\subsubsection{Visual Confounder Dictionary Building}
First, it is necessary to identify the confounder and build the corresponding dictionary. Since the VLN task typically involves indoor environments, the surroundings can be composed of various room types and characteristic objects. Therefore, we mainly consider two observable visual confounders: the \textit{objects} and the \textit{room types}. 

Objects in scenes provide semantic information but can also introduce biases. For example, considering the typical probability distribution of $P(Y|X)=\sum_z P(Y|X,z)P(z|X)$, when $P(z=\texttt{picture}|X=\texttt{wall})$ is large while $P(z=\texttt{door}|X=\texttt{wall})$ is small, the learned features of a \texttt{wall} region are merely its surrounding \texttt{picture}-like features, resulting in biased representations.

Another observable confounder is the room type since the VLN task mainly centers around navigating through different rooms. When certain room types appear frequently in the training set, the model may learn to associate their features, such as connecting the features of the \texttt{bedroom} and \texttt{living room}. As a result, when the agent is given a test trajectory from the \texttt{bedroom} to the \texttt{bathroom}, it may still relate the features of the \texttt{bedroom} with the \texttt{living room} to infer the \texttt{bathroom}, leading to representation bias.

In Fig.~\ref{fig_dict_traj}, we construct the visual confounder dictionaries $D^o_z$ for objects and $D^r_z$ for room types. For the object dictionary, we employ the bottom-up attention model~\cite{anderson2018bottom} to extract the object features for each viewpoint, including their classes, locations, and feature representations. Since the perspective images have substantial overlap, the redundant regions in each panorama are removed as~\cite{majumdar2020improving}. Each entry in $D^o_z$ comprises the average bounding boxes and feature representations corresponding to each object category in the training dataset. Furthermore, since the room types are not initially given in VLN, we propose using a VQA pre-trained model BLIP~\cite{li2022blip} to identify room types by sending images with a fixed prompt ``\textit{what kind of room is this?}" 
After obtaining the room type of each image, CLIP~\cite{radford2021learning} is employed to extract their image-level features, and the average features are used to represent each room type in~$D^r_z$.

\subsubsection{Visual Backdoor Causal Encoder} 
As described in Sec.~\ref{subsec_task_formulation}, each viewpoint consists of several images to present the panorama. Following the previous implementation, the image features $\mathbf{V}=\{\mathbf{v}_i\}_{i=1}^{N}$ and the object features $\mathbf{O}=\{\mathbf{o}_i\}_{i=1}^{M}$ are extracted by CLIP and the bottom-up attention method, respectively. To address the influence of the object confounder and ensure feature alignment, an independent branch is designed to encode object features. Specifically, the self-attention transformer encoders with 2 layers are used to encode the inside relationships of images and objects, respectively. To learn unbiased causal features, the backdoor-based representation learning module proposed in Sec.~\ref{subsec_IBRL} is adopted with the pre-built $\mathcal{D}^o_z$ and $\mathcal{D}^r_z$. After the intervention, a cross-attention transformer encoder is used to align the causal visual features by assigning images as queries and objects as keys and values. The above process is presented as follows:
\begin{align}
    E_v &= \phi_v(\mathbf{V})+\psi^v_p(\gamma_v)+\psi^v_n(N_v) \\
    E_o &= \phi_o(\mathbf{O})+\psi^o_p(B_o)+\psi^o_l(L_o) \\
    C_v &= \Theta(E_v,\mathcal{D}^r_z),\quad C_o = \Theta(E_o,\mathcal{D}^o_z) \\
    Q_v &= \mathcal{T}^s_v(C_v),\quad Q_o = \mathcal{T}^s_o(C_o) \\
    Q_{v,o} &= \mathcal{T}^c_{v,o}(Q_v, Q_o, Q_o)
\end{align}
where $\phi(\cdot)$ and $\psi(\cdot)$ mean the pre-trained encoder and the embedding layer. For image embedding, the orientation information $\gamma_v$ and the navigable types $N_v$ ($1$ for navigable and $0$ for non-navigable) are added to the image features. For object embedding, the bounding box information $B_o$ and the classification logits $L_o$ are added with the object features. $\Theta(\cdot)$ means the backdoor-based representation learning module. $\mathcal{T}^s$ and $\mathcal{T}^c$ denote the self-attention and cross-attention mechanism, respectively.

\subsection{Linguistic Feature Causal Representation Learning}
\label{subsec_LFCR} 
\subsubsection{Linguistic Confounder Dictionary Building}
From a linguistic semantic perspective, both \textit{landmark} and \textit{direction} words play critical roles in guiding navigation instructions. For example, when given an instruction such as ``\textit{Walk past the kitchen and turn right before the stair, then wait at the bedroom}", the landmark words like ``\texttt{kitchen}" and ``\texttt{stair}" provide important references and the direction phrases like ``\texttt{past}" and ``\texttt{right}" guide the agent's forward action. However, similar to the visual inputs, these key phrases can also act as confounders, potentially leading to biased representations.

To address this issue, as shown in Fig.~\ref{fig_dict_instr}, we employ the Natural Language Toolkit (NLTK)~\cite{nltk} to extract relevant phrases and perform direction-landmark recognition based on their part-of-speech tags and entity dictionary. We then use the language encoder BERT~\cite{devlin2018bert} to extract their contextual features. Given BERT's self-attention mechanism, each word may exhibit different features across multiple sentences. To account for this, we compute the average features of each target word across the sentences. In addition, if a word is split into multiple tokens by the BERT tokenizer within a single sentence, we also compute the average features of the split tokens. This process allows us to construct the linguistic confounder dictionaries $\mathcal{D}^w_z$, which includes $\mathcal{D}^d_z$ and $\mathcal{D}^l_z$ for directions and landmarks, respectively.

\subsubsection{Linguistic Backdoor Causal Encoder} To encode the natural language instructions provided as $I=\{w_i\}_{i=1}^{L}$, we employ a BERT model consisting of 9 transformer encoder layers. To enhance sequential information, position encoding $P_w$ is incorporated. Next, the IBRL module $\Theta(\cdot)$ is applied separately to $\mathcal{D}^d_z$ and $\mathcal{D}^l_z$, and their outputs are combined through summation:
\begin{align}
    E_w &= \phi_w(I)+\psi_w(P_w) \\
    U^d_w &= \Theta(E_w, \mathcal{D}^d_z),\quad U^l_w = \Theta(E_w, \mathcal{D}^l_z) \\
    U_w & = U^d_w + U^l_w
\end{align}
where $\phi_w(\cdot)$ and $\psi_w(\cdot)$ denote the BERT encoder and the position embedding, respectively. Furthermore, a gate-like structure incorporating a sigmoid function $\delta(\cdot)$ is employed to dynamically determine the weight value $\omega \in \mathbb{R}^{L\times 1}$, which is used to perform a weighted sum of the linguistic causal features and linguistic context features:
\begin{align}
    \omega &= \delta(U_wW_q+E_wW_e+b_g) \\
    Q_w &= \omega \odot U_w + (1-\omega) \odot E_w
\end{align}
where $W_q\in \mathbb{R}^{d_h\times 1}$, $W_e\in \mathbb{R}^{d_h\times 1}$ and $b_g \in \mathbb{R}^{d_h}$ are learnable parameters. $\odot$ means element-wise multiplication.

\subsection{Memory-Augmented Global-Local Cross-Modal Fusion}
\label{subsec_memory-augmented}
VLN differs from other vision-and-language tasks due to its partially observable Markov decision process, where future actions rely on the current observation. Therefore, it is essential to accurately represent the agent's state. In this paper, we employ the memory-augmented global-local cross-modal fusion proposed by our previous work~\cite{wang2023dual}, consisting of a global adaptive aggregation (GAA) method, a cross-modal encoder, and a recurrent memory fusion~(RMF). GAA computes the visual representation at the $t$-th step as:
\begin{align}
    \mathcal{A} &= \textit{Softmax}(\eta(Q_{v,o}W_a+b_a)) \\
    a_t &= \sum_n^N \mathcal{A} \odot Q_{v,o}
\end{align}
where $\eta(\cdot)$ is the tanh activation function and $N$ presents the number of images. The $t$-th fused visual feature $a_t$ and the features from previous steps form the global graph that captures dependencies among visited nodes. Additionally, a recurrent memory unit is incorporated to enhance reasoning capability. Specifically, the sequences of the global and local branches are built as $G_g=\{\texttt{[CLS]},[a_i]_{i=1}^{t}, \mathcal{R}_{t-1}\}$ and $G_l=\{ \texttt{[CLS]}, Q_{v,o}, \mathcal{R}_{t-1}\}$, respectively. Here, $\mathcal{R}_{t-1}$ means the recurrent memory unit computed in the last step:
\begin{align}
\mathcal{R}_{t-1}=\text{LN}([\mathcal{C}_g,\mathcal{C}_l,\mathcal{C}_w]W_c+b_c)
\end{align}
where LN means layer normalization. $\mathcal{C}_g,\mathcal{C}_l,\mathcal{C}_w$ denote the \texttt{[CLS]} tokens of the global, local and textual branch, respectively. $\mathcal{R}_{t=0}$ is initialized as a zero matrix at the beginning of navigation.
To fuse vision and language features, we use LXMERT~\cite{tan2019lxmert} as the cross-modal encoder, and treat visual features as queries and linguistic features as keys and values:

\begin{align}
    F^l_{v,w} &= \phi^l_f(G_l, Q_w, Q_w) \\
    F^g_{v,w} &= \phi^g_f(G_g, Q_w, Q_w) 
\end{align}
where $\phi^l_f(\cdot)$ and $\phi^g_f(\cdot)$ represent the cross-modal encoders for the local and global branches, respectively.

\subsection{Dynamic Decision-making Strategy}
\label{subsec_decision-making_strategy}
The dynamic decision-making strategy~\cite{chen2022think} is employed to integrate the global and local branches and predict the action. Specifically, $\mathcal{C}_g$ and $\mathcal{C}_l$ are concatenated to compute a one-dimensional scalar weight $\sigma$ using a feed-forward network (FFN). To ensure the consistency of candidate nodes, the local vector $F_{v,w}^l$ is transmitted into the global action space $\hat{F}_{v,w}^l$. Then, two additional FFN networks are used to project the local and global vectors into the scoring domains: 
\begin{align}
    \sigma &= \delta(\phi_c([\mathcal{C}_g,\mathcal{C}_l]))\\
    \mathcal{D}_g &= \phi_g(F_{v,w}^{g}),\quad \hat{\mathcal{D}}_l=\phi_l(\hat{F}^{l}_{v,w})
\end{align}
where $\phi_c,\phi_g$ and $\phi_l$ respectively denote independent FFN networks and $\delta$ means the sigmoid function. 
Lastly, the prediction action $y_t$ is calculated by the weighted sum of two branches, with a mask function $\mathcal{M}(\cdot)$ to filter out non-candidate nodes:
\begin{align}
    y_t &= \textit{Softmax}(\mathcal{M}(\sigma \mathcal{D}_g + (1-\sigma) \hat{\mathcal{D}}_l)) 
\end{align}

\subsection{Training Strategy} 
\label{subsec_training_strategy}
Similar to prior works, our proposed model is trained using the two-step training strategy: pre-training and fine-tuning. 

\subsubsection{Pre-training} 
This paper uses the two most used tasks that are masked language modeling (MLM)~\cite{devlin2018bert} and single-step action prediction (SAP)~\cite{chen2021history} for pre-training R2R and RxR. Additionally, object grounding (OG)~\cite{Lin_2021_CVPR} is used for REVERIE. Specifically, MLM requires the model to predict masked words $w_m$ based on surrounding words $w_{\backslash m}$ and the visual trajectory $\tau$. SAP and OG are off-line expert demonstrations with behavior cloning, which predict the action at the $t$-th step and the object for the given instruction $I$:
\begin{align}
    \label{eq_loss_mlm}
    \mathcal{L}_{\text{MLM}} &= -\log f_\theta(w_m|w_{\backslash m},\tau) \\
    \label{eq_loss_sap}
    \mathcal{L}_{\text{SAP}} &= - \sum_{t=1}^{T} \log f_\theta(\hat{a}_t|I,\hat{\tau}_{1:t-1},v_t) \\
    \label{eq_loss_og}
    \mathcal{L}_{\text{OG}} &= -\log f_\theta(\hat{o}|I,\tau)
\end{align}
where $\hat{a}_t$, $v_t$, and $\hat{o}$ mean the $t$-th ground-truth action, visual observation, and the target object, respectively. 


\subsubsection{Fine-tuning} To balance accuracy and robustness in fine-tuning the sequence-to-sequence VLN task, we employ a combination of teacher-forcing and sampling strategies. The teacher-forcing strategy uses ground-truth input as supervision, while the sampling strategy is based on the pseudo interactive demonstrator (PID)~\cite{chen2022think} to select the nearest navigable node to the final destination in each iteration. The decision-making loss is calculated using cross-entropy loss.
\begin{align}
    \mathcal{L}_{\text{FT}} &= -\sum_{t=1}^{T} \log f_\theta(\hat{a}_t|I,\tau_{1:t-1},v_t)
\end{align}

Drawing on the success of our previous works~\cite{wang2023res,wang2023pasts}, we also explored the effect of data augmentation on CausalVLN. Specifically, we utilize PASTS~\cite{wang2023pasts} on R2R and RES-StS~\cite{wang2023res} on REVERIE to generate pseudo labels using the pre-trained speaker model with the environmental dropout. 

\section{Experiments}
\label{sec_experiments}
\begin{table*}[!htp]
\centering
\caption{Comparison with the state-of-the-art methods on the R2R dataset. * means using PASTS~\cite{wang2023pasts} during the fine-tuning stage.}
\begin{tabular}{l|cccc|cccc|cccc}
\toprule
\multirow{2}{*}{Method} & \multicolumn{4}{c|}{R2R Validation Seen} & \multicolumn{4}{c|}{R2R Validation Unseen} & \multicolumn{4}{c}{R2R Test Unseen} \\
 & SR$\uparrow$ & SPL$\uparrow$ & NE$\downarrow$ & OSR$\uparrow$ & SR$\uparrow$ & SPL$\uparrow$ & NE$\downarrow$ & OSR$\uparrow$ & SR$\uparrow$ & SPL$\uparrow$ & NE$\downarrow$ & OSR$\uparrow$ \\ \midrule
EnvDrop~\cite{tan2019learning} & 62 & 59 & 3.99 & - & 52 & 48 & 5.22 & - & 51 & 47 & 5.23 & 59 \\
RCM~\cite{wang2020vision} & 67 & - & 3.53 & - & 43 & - & 6.09 & - & 43 & 38 & 6.12 & - \\
AuxRN~\cite{zhu2020vision} & 70 & 67 & 3.33 & 78 & 55 & 50 & 5.28 & 62 & 55 & 51 & 5.15 & 62 \\
NvEM~\cite{an2021neighbor} & 69 & 65 & 3.44 & - & 60 & 55 & 4.27 & - & 58 & 54 & 4.37 & 66 \\ \hline
PREVALENT~\cite{hao2020towards} & 60 & 65 & 3.67 & - & 57 & 53 & 4.73 & - & 54 & 51 & 4.75 & 61 \\
RecBERT~\cite{hong2021vln} & 72 & 68 & 2.90 & 79 & 63 & 57 & 3.93 & 69 & 63 & 57 & 4.09 & 70 \\
HAMT~\cite{chen2021history} & 76 & 72 & 2.51 & 82 & 66 & 61 & 3.29 & 73 & 65 & 60 & 3.93 & 72 \\
REM~\cite{liu2021vision} & 75 & 72 & 2.48 & - & 64 & 58 & 3.89 & - & 65 & 59 & 3.87 & - \\
HOP+~\cite{qiao2023hop_plus} & 78 & 73 & 2.33 & - & 67 & 61 & 3.49 & - & 66 & 60 & 3.71 & - \\
DUET~\cite{chen2022think} & 79 & 73 & 2.28 & 86 & 72 & 60 & 3.31 & 81 & 69 & 59 & 3.65 & 76 \\
DSRG~\cite{wang2023dual} & 81 & 76 & 2.23 & 88 & 73 & 62 & 3.00 & 81 & 72 & 61 & 3.33 & 78 \\
\textbf{CausalVLN} & 79 & 73 & 2.42 & 85 & 74 & 64 & 2.83 & 82 & 72 & 62 & 3.24 & 79 \\
\hline
EnvDrop*~\cite{tan2019learning} & 69 & 65 & 3.38 & - & 55 & 50 & 4.80 & - & 56 & 52 & 5.30 & - \\
RecBERT*~\cite{hong2021vln} & 72 & 67 & 3.13 & 78 & 63 & 58 & 3.99 & 69 & 64 & 59 & 4.07 & 70 \\
HAMT*~\cite{chen2021history} & 77 & 74 & 2.46 & 83 & 68 & 62 & 3.37 & 76 & 67 & 61 & 3.77 & 73 \\
DUET*~\cite{chen2022think} & 82 & \textbf{79} & \textbf{1.89} & 87 & 72 & 62 & 3.01 & 80 & 70 & 60 & 3.55 & 76 \\
DSRG*~\cite{wang2023dual} & 81 & 76 & 1.92 & 87 & 74 & 64 & 2.77 & 82 & - & - & - & - \\
\textbf{CausalVLN*} & \textbf{82} & 77 & 1.92 & \textbf{88} & \textbf{76} & \textbf{67} & \textbf{2.67} & \textbf{84} & \textbf{73} & \textbf{64} & \textbf{3.15} & \textbf{79} \\ \bottomrule
\end{tabular}
\label{tab_r2r}
\end{table*}

\subsection{Datasets}
The proposed CausalVLN model is evaluated on three VLN datasets: VLN with fine-grained instructions (R2R~\cite{anderson2018vision} and RxR~\cite{ku2020room}), and VLN with high-level instructions (REVERIE~\cite{qi2020reverie}). The datasets are constructed based on the Matterport3D simulator~\cite{chang2017matterport3d} and are divided into four subsets: the training set, the validation-seen set (with the same buildings as the training set), the validation-unseen set (with new buildings), and the test-unseen set. To assess the model's generalization and robustness, we place particular emphasis on evaluating its performance in unseen environments.

\textbf{Room-to-Room (R2R)} This dataset encompasses images from 90 distinct buildings, comprising 21,576 navigation instructions with an average length of 29 words, and 7,189 paths. The dataset contains instructions that incorporate detailed linguistic cues, such as \textit{``Enter the room on the left. Turn slightly left and wait at the door."}

\textbf{Room-across-Room (RxR)} RxR enhances R2R by offering a larger dataset, addressing path biases, and eliciting more references to visible entities. 
We use the English subset of RxR (en-IN and en-US), which consists of 42,002 instructions, with an average length of 108 words.

\textbf{REVERIE} This dataset emphasizes high-level instructions that involve both reaching a goal destination and finding a specific target object. It contains 21,702 instructions, with an average length of 18 words. The instructions are concise, such as \textit{``Please go to the kitchen and clean the sink."}


\subsection{Implementation Details}
We adopt a similar configuration to DSRG~\cite{wang2023dual}, with 9 layers for the language encoder, 2 layers for the vision encoder, and 4 layers for the cross-modal encoder. To intervene with the object confounders, we use the object features extracted by the bottom-up attention model provided by~\cite{majumdar2020improving}. The object classes are normalized to 43 based on the category map from Matterport3D~\cite{chang2017matterport3d}. 
Following~\cite{shen2021much}, CLIP-ViT-B~\cite{radford2021learning} is used to extract image features. The top 50 room types with the highest occurrence output by BLIP~\cite{li2022blip} are calculated. During the pre-training phase, our CausalVLN is trained on a single Tesla V100 GPU with a batch size of 24 for 200K iterations. The AdamW optimizer~\cite{loshchilov2017decoupled} with a learning rate of $5\times 10^{-5}$ is employed. 
Marky-mT5~\cite{wang2022less} is used for RxR data augmentation. In the fine-tuning stage, we employ batch size 8 for R2R and REVERIE, and 5 for RxR. The learning rate is set to $1\times 10^{-5}$, and the maximum number of iterations is 100K. 
Additionally, when employing PASTS~\cite{wang2023pasts} for R2R and RES-StS~\cite{wang2023res} for REVERIE based on back translation~\cite{tan2019learning}, the environmental dropout is set to 0.5. We alternate between annotated data and dynamically generated speaker-enhanced data for each iteration. The best model is selected based on its performance on the validation unseen split, with the optimal checkpoint saved for R2R based on SPL, for RxR based on sDTW, and for REVERIE based on RGS.

\begin{table*}[!htp]
\centering
\caption{Comparison with the state-of-the-art methods on the REVERIE dataset. * means using RES-StS~\cite{wang2023res} during the fine-tuning stage.}
\resizebox{\linewidth}{!}{
\renewcommand{\arraystretch}{1.2}
\begin{tabular}{l|ccccc|ccccc|ccccc}
\toprule
\multirow{3}{*}{Methods} & \multicolumn{5}{c|}{REVERIE Validation Seen} & \multicolumn{5}{c|}{REVERIE Validation Unseen} & \multicolumn{5}{c}{REVERIE Test Unseen} \\
 & \multicolumn{3}{c}{Navigation} & \multicolumn{2}{c|}{Grounding} & \multicolumn{3}{c}{Navigation} & \multicolumn{2}{c|}{Grounding} & \multicolumn{3}{c}{Navigation} & \multicolumn{2}{c}{Grounding} \\
 & OSR$\uparrow$ & SR$\uparrow$ & SPL$\uparrow$ & RGS$\uparrow$ & RGSPL$\uparrow$ & OSR$\uparrow$ & SR$\uparrow$ & SPL$\uparrow$ & RGS$\uparrow$ & RGSPL$\uparrow$ & OSR$\uparrow$ & SR$\uparrow$ & SPL$\uparrow$ & RGS$\uparrow$ & RGSPL$\uparrow$ \\ \midrule
Seq2Seq~\cite{anderson2018vision} & 35.70 & 29.59 & 24.01 & 18.97 & 14.96 & 8.07 & 4.20 & 2.84 & 2.16 & 1.63 & 6.88 & 3.99 & 3.09 & 2.00 & 1.58 \\
RCM~\cite{wang2020vision} & 29.44 & 23.33 & 21.82 & 16.23 & 15.36 & 14.23 & 9.29 & 6.97 & 4.89 & 3.89 & 11.68 & 7.84 & 6.67 & 3.67 & 3.14 \\
SIA~\cite{Lin_2021_CVPR} & 65.85 & 61.91 & 57.08 & 45.96 & 42.65 & 44.67 & 31.53 & 16.28 & 22.41 & 11.56 & 44.56 & 30.80 & 14.85 & 19.02 & 9.20 \\
RecBERT~\cite{hong2021vln} & 53.90 & 41.79 & 47.96 & 38.23 & 35.61 & 35.02 & 30.67 & 24.90 & 18.77 & 15.27 & 32.91 & 29.61 & 23.99 & 16.50 & 13.51 \\
HOP~\cite{qiao2022hop} & 54.88 & 53.74 & 47.19 & 38.65 & 33.85 & 36.24 & 31.78 & 26.11 & 18.85 & 15.73 & 33.06 & 30.17 & 24.34 & 17.69 & 14.34 \\
HAMT~\cite{chen2021history} & 47.65 & 43.29 & 40.19 & 27.20 & 25.18 & 36.84 & 32.95 & 30.20 & 18.92 & 17.28 & 33.41 & 30.40 & 26.67 & 14.88 & 13.08 \\
HOP+~\cite{qiao2023hop_plus} & 56.43 & 55.87 & 49.55 & 40.76 & 36.22 & 40.04 & 36.07 & 31.13 & 22.49 & 19.33 & 35.81 & 33.82 & 28.24 & 20.20 & 16.86 \\
DUET~\cite{chen2022think} & 73.86 & 71.75 & 63.94 & 57.41 & 51.14 & 51.07 & 46.98 & 33.73 & 32.15 & 23.03 & 56.91 & 52.51 & 36.06 & 31.88 & 22.06 \\
DSRG~\cite{wang2023dual} & 77.72 & 75.69 & 68.09 & 61.07 & 54.72 & 53.25 & 47.83 & \textbf{34.02} & 32.69 & \textbf{23.37} & 58.26 & 54.04 & 37.09 & 32.49 & 22.18 \\
\textbf{CausalVLN} & 79.34 & 77.51 & 69.59 & 62.33 & 55.99 & 57.80 & 50.44 & 30.41 & 34.82 & 20.72 & 62.51 & 55.93 & 37.18 & 33.47 & 22.34 \\ \hline
HAMT* & 60.65 & 58.54 & 56.01 & 42.66 & 40.70 & 37.09 & 34.25 & 30.31 & 20.48 & 18.09 & 39.48 & 37.38 & 32.65 & 20.07 & 17.50 \\
DUET* & 78.50 & 75.40 & 67.13 & 62.08 & 55.39 & 55.01 & 48.85 & 33.07 & 33.17 & 22.33 & 62.41 & 57.23 & \textbf{38.61} & 35.33 & \textbf{23.64} \\
\textbf{CausalVLN*} & \textbf{82.78} & \textbf{80.46} & \textbf{70.69} & \textbf{64.93} & \textbf{57.16} & \textbf{58.79} & \textbf{51.26} & 32.43 & \textbf{35.39} & 22.54 & \textbf{66.83} & \textbf{59.30} & 38.59 & \textbf{35.55} & 23.11 \\ \bottomrule
\end{tabular}}
\label{tab_reverie}
\end{table*}

\subsection{Quantitative Results}
\subsubsection{Room-to-Room (R2R)}

\textbf{Evaluation Metrics}
Five metrics are used to evaluate: \texttt{Navigation Error (NE)} quantifies the distance between the predicted stop location and the reference location. \texttt{Success Rate (SR)} measures the frequency at which the predicted stop location falls within a certain threshold distance of the ground truth. 
\texttt{Oracle Success Rate (OSR)} measures proximity to the goal by assessing how often any node in the predicted path falls within a threshold distance of the target location. \texttt{The Success Rate weighted by Inverse Path Length (SPL)} evaluates navigation performance by considering both the success rate and the efficiency of the navigation process.

\textbf{Comparison with SoTA}
Table~\ref{tab_r2r} showcases the superior performance of our proposed CausalVLN model compared to state-of-the-art approaches on the R2R dataset. CausalVLN achieves the best results across all metrics on the validation unseen set. Incorporating fine-tuning with PASTS~\cite{wang2023pasts} further enhances navigation performance, with notable improvements in SR ($\uparrow 3\%$) and SPL ($\uparrow 5\%$) compared to DSRG. These enhancements are also observed on the test unseen set, solidifying the effectiveness of our CausalVLN model.

Although a slight decline is observed in performance on the seen validation set, we believe this can be attributed to the inherent bias of the R2R dataset towards shortest paths and its limited size~\cite{anderson2018vision}. Overfitting to specific routes during training can lead to a biased understanding of the environment and hinder generalization to new environments. By incorporating causal intervention in feature learning, we mitigate biases stemming from spurious correlation in path selection. Consequently, the decline in performance in seen environments is accompanied by an improvement in new environments. Additionally, using PASTS for additional training paths enhances performance in both seen and unseen environments and showcases the significant potential for CausalVLN to synergize with data augmentation methods.

\subsubsection{REVERIE}

\textbf{Evaluation Metrics} Since REVERIE requires the agent to identify an object at the target location, two additional metrics are used: \texttt{Remote Grounding Success rate (RGS)} measures the correct object grounding ratio, and \texttt{RGS weighted by Path Length (RGSPL)} consider both RGS and the navigation length.

\textbf{Comparison with SoTA}
In Table~\ref{tab_reverie}, we compare our CausalVLN with other methods on the REVERIE dataset. Our model shows superior performance across multiple metrics. Notably, compared with the previous state-of-the-art DSRG~\cite{wang2023dual}, CausalVLN gains significant improvements in SR ($\uparrow 2.61\%$) and RGS ($\uparrow 2.13\%$) on the validation unseen set. Similar enhancements are also observed on the validation seen and test unseen sets. Additionally, by incorporating RES-StS~\cite{wang2023res} for data augmentation, CausalVLN achieves better performance on most metrics, improving SR by $2.41\%$ and RGS by $2.22\%$. This highlights the effectiveness of our method in tackling the high-level VLN task. However, we do observe a decline in SPL and RGSPL in unseen environments. We interpret this as a result of CausalVLN's tendency to explore more in unseen environments, which leads to higher SR and RGS. Moreover, given the lack of specific route guidance in high-level instructions, it is possible for the agent to reach the target goal without strictly following the ground-truth path. Moving forward, we remain committed to addressing these challenges and improving navigation efficiency in unseen environments as part of our future work.
\begin{table}[tb]
\setlength\tabcolsep{4pt}
\centering
\caption{Comparison with the state-of-the-art methods on the RxR English dataset.}
\large
\resizebox{\linewidth}{!}{
\renewcommand{\arraystretch}{1.2}
\begin{tabular}{l|cccc|cccc}
\toprule
\multirow{2}{*}{Method} & \multicolumn{4}{c|}{RxR Validation Seen} & \multicolumn{4}{c}{RxR Validation Unseen} \\
 & SR$\uparrow$ & SPL$\uparrow$ & nDTW$\uparrow$ & sDTW$\uparrow$ & SR$\uparrow$ & SPL$\uparrow$ & nDTW$\uparrow$ & sDTW$\uparrow$ \\ \bottomrule
Baseline~\cite{ku2020room} & 28.6 & - & 45.4 & 23.2 & 26.1 & - & 42.4 & 21.0 \\
EnvDrop~\cite{tan2019learning} & 48.1 & 44.0 & 57.0 & 40.0 & 38.5 & 34.0 & 51.0 & 32.0 \\
Syntax~\cite{li2021improving} & 48.1 & 44.0 & 58.0 & 40.0 & 39.2 & 35.0 & 52.0 & 32.0 \\
HOP~\cite{qiao2022hop} & 49.4 & 45.3 & 58.0 & 40.0 & 42.3 & 36.3 & 52.0 & 33.0 \\
SOAT~\cite{moudgil2021soat} & - & - & - & - & 44.2 & - & 54.8 & 36.4 \\
HOP+~\cite{qiao2023hop_plus} & 53.6 & 47.9 & 59.0 & 43.0 & 45.7 & 38.4 & 52.0 & 36.0 \\
FOAM~\cite{dou2022foam} & - & - & - & - & 42.8 & 38.7 & 54.1 & 35.6 \\
ADAPT~\cite{lin2022adapt} & 50.3 & 44.6 & 56.3 & 40.6 & 46.9 & 40.2 & 54.1 & 37.7 \\
$\text{MARVAL}_{\text{M-MP}}$~\cite{kamath2023new} & - & - & - & - & 50.2 & - & 60.3 & 43.9 \\ 
\textbf{CausalVLN} & \textbf{68.6} & \textbf{62.7} & \textbf{66.6} & \textbf{56.1} & \textbf{62.0} & \textbf{54.2} & \textbf{61.3} & \textbf{49.4} \\ \bottomrule
\end{tabular}}
\label{tab_rxr}
\end{table}

\subsubsection{Room-across-Room (RxR)}

\textbf{Evaluation Metrics} 
Two additional metrics are employed: \texttt{normalized Dynamic Time Warping (nDTW)}~\cite{ilharco2019general}, which provides the alignment between the predicted path and the ground truth path, and \texttt{success rate weighted by Dynamic Time Warping (sDTW)}, which captures both the successful completion of reaching the target destination and the accuracy of the predicted trajectory.

\textbf{Comparison with SoTA} Table~\ref{tab_rxr} presents the experimental results on the RxR English dataset, showing the superior performance of our CausalVLN model compared to previous approaches. Notably, when compared to MARVAL~\cite{kamath2023new}, which also utilizes Marky-generated instructions in Matterport3D (M-MP), our method demonstrates substantial improvements in SR and sDTW of 11.8\% and 5.5\%, respectively, in unseen environments. These remarkable enhancements underscore the effectiveness and robustness of our proposed intervention method, establishing a strong foundation for future research endeavors.

\subsection{Ablation Study}
\subsubsection{Effect of the Intervention}
\begin{figure}[tb]
    \centering
    \includegraphics[scale=0.6]{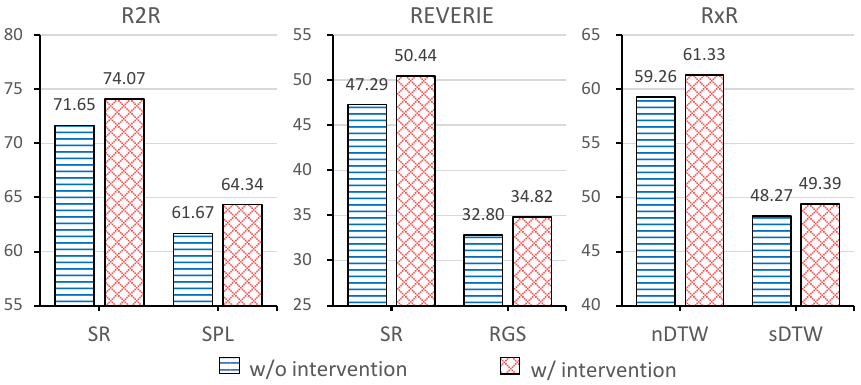}
    \caption{The effects of the intervention on R2R, REVERIE and RxR datasets.}
    \label{fig_intervention_multidataset}
\end{figure}
\begin{table}[tb]
\centering
\caption{Effect of different Confounders.}
\renewcommand{\arraystretch}{1.2}
\begin{tabular}{c|ccc|cccc}
\toprule
Idx & Instr & Obj & Room & SR$\uparrow$ & SPL$\uparrow$ & NE$\downarrow$ & OSR$\uparrow$ \\ \midrule
\#1 &  &  &  & 71.65 & 61.67 & 3.32 & 79.05 \\ \hline
\#2 & \checkmark &  &  & 72.46 & 61.94 & 3.13 & 80.33 \\
\#3 &  & \checkmark &  & 72.29 & 63.08 & 3.17 & 80.59 \\
\#4 &  &  & \checkmark & 72.71 & 63.43 & 3.06 & 80.12 \\ \hline
\#5 &  & \checkmark & \checkmark & 73.18 & 63.37 & 3.11 & 80.25 \\
\#6 & \checkmark & \checkmark &  & 72.12 & 62.09 & 3.21 & 80.20 \\
\#7 & \checkmark &  & \checkmark & 74.05 & 63.67 & 2.89 & \textbf{83.14} \\ \hline
\#8 & \checkmark & \checkmark & \checkmark & \textbf{74.07} & \textbf{64.34} & \textbf{2.83} & 81.69 \\ \bottomrule
\end{tabular}
\label{tab_confounder_factors}
\end{table}
In Fig.~\ref{fig_intervention_multidataset}, we present the effects of the intervention on the R2R, REVERIE, and RxR validation unseen set, where the most important metrics for each dataset are illustrated. It can be found that with the help of causal learning on the feature representation through the intervention process, the model performance in unseen environments has got significant improvements. For R2R, the improvements of SR ($\uparrow 2.42\%$) and SPL ($\uparrow 2.67\%$) indicate more effective and efficient achievement of the correct target location based on instructions. The enhancement of RGS ($\uparrow 2.02\%$) for REVERIE suggests improved grounding of remote objects during navigation. The similar improvements in RxR, with the improvements of $2.07\%$ in nDTW and $1.12\%$ in sDTW, also indicate that the intervention can make the agent better understand long instructions and follow the trajectory.

\subsubsection{Effect of Different Confounders}
In Table~\ref{tab_confounder_factors}, we investigate the impact of different confounders on the R2R validation unseen set. As discussed in Sec.~\ref{subsec_VFCR} and Sec.~\ref{subsec_LFCR}, the considered observable confounders include the direction-landmark tokens in instructions (denoted as \textit{Instr}) and the objects (denoted as \textit{Obj}) and room types (denoted as \textit{Room}) in visual observations. From \#2 to \#4, it can be observed that each type of confounder individually contributes to performance improvements compared to the baseline (\#1) where no interventions are made. Furthermore, \#5 to \#7 indicate that by reasonably increasing the intervention of inputs, the model can better learn effective causal representations. For example, when both instructions and room types are intervened, compared to only intervened instructions, SR ($\uparrow 1.59\%$) and SPL ($\uparrow 1.72\%$) obtain significant improvement. 
\#8 demonstrates that most metrics can reach the best results when all confounders are considered together. This strongly proves the reliability and rationality of our hypothesis about observable confounders hidden in language and vision for VLN. By using the backdoor adjustment method, the model acquires a more holistic understanding of environments and instructions, resulting in improved navigation capabilities.

\subsubsection{Effect of Intervention Methods for Vision and Language}
\begin{table}[tb]
\centering
\caption{Comparison of Intervention Methods for Vision and Language.}
\begin{tabular}{cc|cccc}
\toprule
Language & Vision & SR$\uparrow$ & SPL$\uparrow$ & NE$\downarrow$ & OSR$\uparrow$ \\ \midrule
Type-1 & Type-1 & 72.54 & 63.73 & 3.10 & 80.72 \\
Type-2 & Type-2 & 71.90 & 61.94 & 3.23 & 81.01 \\
Type-1 & Type-2 & 71.43 & 62.47 & 3.24 & 78.33 \\
\textbf{Type-2} & \textbf{Type-1} & \textbf{74.07} & \textbf{64.34} & \textbf{2.83} & \textbf{81.69} \\ 
\bottomrule
\end{tabular}
\label{tab_IBRL_design}
\end{table}

\begin{table}[tb]
\centering
\caption{Comparison of Iterative Strategies in IBRL.}
\renewcommand{\arraystretch}{1.2}
\begin{tabular}{cl|cccc}
\toprule
\multicolumn{2}{c|}{Method} & SR$\uparrow$ & SPL$\uparrow$ & NE$\downarrow$ & OSR$\uparrow$ \\ \midrule
\multicolumn{2}{c|}{Random} & 72.24 & 63.14 & 3.20 & 81.01 \\
\multicolumn{2}{c|}{Pre-computing} & 72.92 & 63.42 & 3.03 & 81.74 \\ \hline
\multicolumn{1}{c|}{\multirow{3}{*}{Iterative}} & Schedule only & 72.88 & 63.59 & 2.96 & 81.01 \\
\multicolumn{1}{c|}{} & Best only & 74.03 & 63.69 & 2.96 & \textbf{82.97} \\
\multicolumn{1}{c|}{} & Best + Schedule & \textbf{74.07} & \textbf{64.34} & \textbf{2.83} & 81.69 \\ \bottomrule
\end{tabular}
\label{tab_iterative_strategy}
\end{table}

As discussed in Sec.~\ref{subsec_IBRL}, we have designed two different intervention methods for learning causal features from inputs. Type-1 denotes the statistic-based model where the probability of each confounder is obtained from the dataset. Type-2 represents the attention-based method, where the probability is calculated through the attention mechanism between inputs and confounders. Table~\ref{tab_IBRL_design} displays the results on the R2R validation unseen set, indicating that the optimal performance is attained when employing the attention-based method for language and the statistic-based method for vision.

The findings suggest that the effects of different intervention models vary across modalities and scenarios. Language inputs are well-suited for the adaptive attention-based intervention method due to their highly structured contextual relevance. Additionally, the BERT architecture used for encoding text embeddings undergoes end-to-end training, allowing the attention-based method to align confounders more accurately with different text inputs. Conversely, image features have less obvious structural and contextual information, and the pre-trained image extractor used for pre-processing is not involved in training. Consequently, employing attention-based intervention may unintentionally interfere with the representations of original image features. In such cases, employing a more stable statistic-based method can provide better treatment, thereby enhancing the performance of feature representation.

\subsubsection{Effect of Iterative Strategies in IBRL}
Table~\ref{tab_iterative_strategy} compares the effects of different strategies for initializing and updating the confounder features in our proposed IBRL. The strategies include \textit{Random} initialization, \textit{Pre-computing} with fixed average features, and \textit{Iterative} updating during training. The iterative approach includes two methods: \textit{Schedule}, which follows a fixed update schedule (default is set to 3K iterations), and \textit{Best}, which updates features when the model achieves improved results during training. The results demonstrate that utilizing pre-trained models to extract confounder features leads to improved performance compared to random initialization. However, as the training progresses, a misalignment can occur between the pre-calculated features and the updated features. To overcome this limitation, we propose an iterative approach that enables timely updates of the confounder features. Experimental results show that the iterative approach, which continuously adapts the representations of the confounder features, significantly enhances intervention performance compared to other methods.

\begin{figure}[tb]
    \centering
    \includegraphics[scale=0.4]{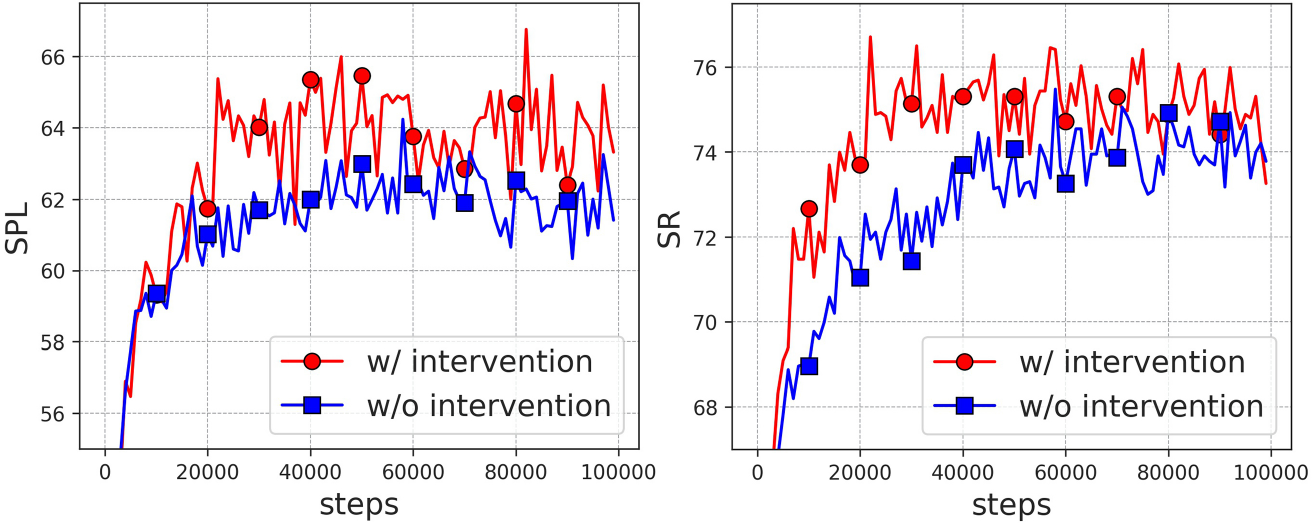}
    \caption{The learning curves of SPL and SR with and without the intervention.}
    \label{fig_curves}
\end{figure}
\begin{figure}[tb]
    \centering
    \includegraphics[scale=0.7]{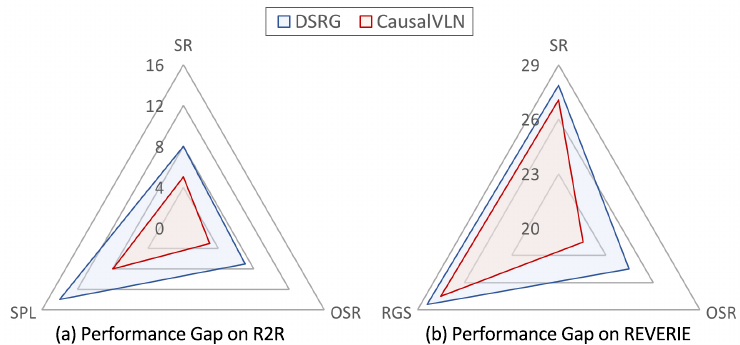}
    \caption{Comparison of performance gaps on R2R and REVERIE between seen and unseen environments. Smaller gaps indicate better generalization.}
    \label{fig_gap}
\end{figure}
\subsection{Qualitative Results}
\subsubsection{Visualization of Learning Curves}
Fig.~\ref{fig_curves} provides the learning curves obtained from the R2R validation unseen set, comparing models trained with and without intervention under the same settings. The visualization reveals that the model incorporating intervention exhibits faster convergence and superior performance in both SR and SPL compared to the model without intervention. This observation clearly demonstrates the significant impact of the proposed causal feature representation learning in enhancing the robustness and generalization performance of the model.

\subsubsection{Comparison of Performance Gaps between Seen and Unseen Environments}
Fig.~\ref{fig_gap} presents the comparison of performance gaps between seen and unseen environments on the R2R and REVERIE datasets, which are known to be highly biased in paths~\cite{ku2020room,zhang2021diagnosing}. Compared with DSRG~\cite{wang2023dual}, it shows that our CausalVLN successfully narrows down the performance gaps on the key metrics for different datasets. These results demonstrate the effectiveness of CausalVLN in mitigating perceptual and reasoning biases, leading to improved generalization and robustness across various environments.

\subsubsection{Visualization of Navigation Trajectory}
\begin{figure*}[!htb]
    \centering
    \includegraphics[scale=0.65]{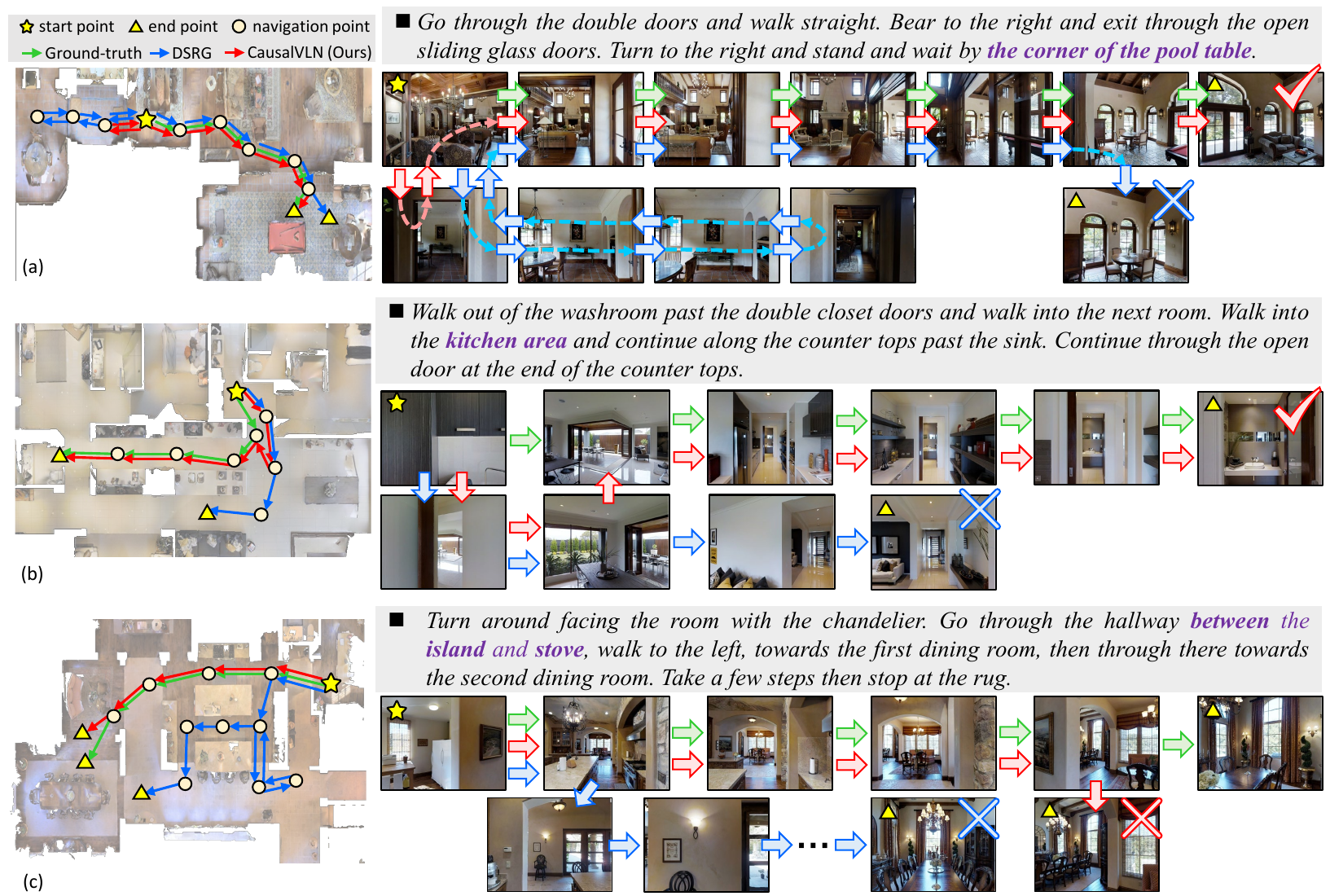}
    \caption{The comparisons of predicted trajectories. The figure showcases the top views of the navigation routes on the left, accompanied by the instructions and the corresponding first-person perspective views on the right. The ground truth is depicted in green, while the predicted results of DSRG~\cite{wang2023dual} and the proposed CausalVLN are represented in blue and red, respectively. Key elements that contribute to different outcomes are highlighted in bold purple characters.}
    \label{fig_traj_visualize}
\end{figure*}
Fig.~\ref{fig_traj_visualize} presents visualizations of trajectories on the R2R validation unseen set, comparing the predictions made by the proposed CausalVLN with those of DSRG and the ground truth results. In Fig.~\ref{fig_traj_visualize} (a), it is evident that CausalVLN exhibits superior error correction capability, accurately identifying key target descriptions such as \textit{``the corner of the pool table"}, while DSRG fails and stops at the wrong location. Similarly, in Fig.~\ref{fig_traj_visualize} (b), CausalVLN successfully predicts the route to the kitchen, whereas DSRG overshoots and proceeds into the living room. Fig.~\ref{fig_traj_visualize} (c) presents a failure case for CausalVLN, where the stop location is at the rug in the first dining room instead of the second dining room as instructed. This suggests that the model is less sensitive to numerical descriptions, which inspires our future work. Nevertheless, upon analyzing the entire route, it is evident that CausalVLN accurately captures the path \textit{``between the island and stock"}, while avoiding significant deviations observed in predictions by DSRG. Given that the models are tested in unseen environments, these results indicate that the learned causal representations enable CausalVLN to accurately reason the paths it has not encountered previously.

\section{Conclusion}
\label{sec_conclusion}
In this work, we present CausalVLN, a novel framework to address the challenge of weak generalization caused by the spurious correlation in the Vision-and-Language Navigation (VLN) task. Specifically, we leverage a causal learning paradigm to train a more robust navigator capable of learning unbiased linguistic and visual representations. First, we employ the structural causal model (SCM) to establish reasonable assumptions about the observable confounders in VLN. Based on the characteristics of the VLN, we consider direction-landmark tokens as confounders for linguistic instructions, and objects and room types as confounders for visual observations. Then, we propose an iterative backdoor-based representation learning (IBRL) method, which enables flexible and adaptive intervention on confounders and inputs. Next, we introduce a visual backdoor causal encoder and a linguistic backdoor causal encoder, enabling the realization of unbiased feature representations during training and validation. Through extensive experiments on three widely used VLN datasets (R2R, RxR, and REVERIE), our proposed method consistently outperforms state-of-the-art approaches, demonstrating its superiority. Furthermore, the visualization analysis validates the effectiveness of CausalVLN in narrowing down the performance gap between seen and unseen environments, demonstrating its strong generalization capability. This research provides valuable insights to achieve robust and generalized VLN agents, and also paves the way for advancements in other related fields by unraveling the potential of causal learning in overcoming biases and improving feature representation.

\section*{Acknowledgments}
This paper is supported by the National Natural Science Foundation of China under Grants (62233013, 62073245, 62173248). 
\bibliographystyle{IEEEtran}
\bibliography{main}

\end{document}